\pdfoutput=1
\documentclass[11pt]{article}

\usepackage[]{EACL2023}
\usepackage{times}
\usepackage{latexsym}
\usepackage{booktabs}
\usepackage{graphicx}
\usepackage{amsfonts} 
\usepackage{amsmath}
\usepackage{xspace}
\newcommand{\modelname}{QAMAT\xspace}
\newcommand{\modelnamemultihop}{QAMAT+\xspace}

\usepackage[T1]{fontenc}
\usepackage[utf8]{inputenc}

\usepackage{microtype}
\usepackage{amsmath}
\usepackage{inconsolata}

\title{Augmenting Pre-trained Language Models with \\QA-Memory for Open-Domain Question Answering}

\author{Wenhu Chen, Pat Verga, Michiel de Jong$^\dagger$, John Wieting, William W. Cohen \\
  Google Research, University of Southern California$^\dagger$  \\
  \small \texttt{\{wenhuchen,patverga,jwieting,wcohen\}@google.com, msdejong@usc.edu}
  \\}

\begin{document}
\maketitle
\begin{abstract}
Existing state-of-the-art methods for open-domain question-answering (ODQA) use an open book approach in which information is first retrieved from a large text corpus or knowledge base (KB) and then reasoned over to produce an answer. A recent alternative is to retrieve from a collection of previously-generated question-answer pairs; this has several practical advantages including being more memory and compute-efficient. Question-answer pairs are also appealing in that they can be viewed as an intermediate between text and KB triples: like KB triples, they often concisely express a single relationship, but like text, have much higher coverage than traditional KBs. In this work, we describe a new QA system that augments a text-to-text model with a large memory of question-answer pairs, and a new pre-training task for the latent step of question retrieval. The pre-training task substantially simplifies training and greatly improves performance on smaller QA benchmarks. Unlike prior systems of this sort, our QA system can also answer multi-hop questions that do not explicitly appear in the collection of stored question-answer pairs.
\end{abstract}

\section{Introduction \label{sec:intro}}

Open-domain question answering (ODQA) is a well-studied knowledge-intensive task. State-of-the-art methods require retrieving relevant knowledge from a large corpus or datastore before reasoning over this retrieved evidence. Most existing methods retrieve documents~\cite{chen2017reading,lee2019latent,karpukhin2020dense} or structured KB triples~\cite{sun2021adaptable}. Recently, a few works have proposed retrieving from a collection of question-answer (QA) pairs---an approach made feasible by advances in scalable automatic question generation.  In this setting, a new question is answered by retrieving paraphrases from a question index, and returning the associated answer \citep{xiao2021open, lewis2021paq}. Notably, the RePAQ system from \citep{lewis2021paq} won the 2020 EfficientQA competition~\cite{min2021neurips}, outperforming closed-book QA (CBQA) models by a significant margin and matching the prior SoTA performance on NQ~\cite{kwiatkowski2019natural}.

A collection of QA pairs is appealing for several reasons.  As opposed to text passages and much like a KB triple, QA pairs are often concise, tending to express a single relationship. However, unlike KB triples, QA collections have good coverage of actually asked questions like those in standard open QA datasets. 
% RePAQ demonstrated several advantageous properties:  Relative to common text retrieval methods, the RePAQ approach is memory and computationally efficient. It is an accurate ``selective QA system''---i.e., where precision can be increased by selectively abstaining, instead of returning a low-confidence answer. Finally, RePAQ systems can be effectively ensembled with text-retrieval QA systems, leading to a hybrid system that outperforms either.
RePAQ demonstrated several advantageous properties such as memory and computational efficiency, strong selective QA performance (i.e. selectively abstaining from answering), and effective ensembling with text-retrieval QA systems.

\begin{figure*}[!t]
    \centering
    \includegraphics[width=1.0\linewidth]{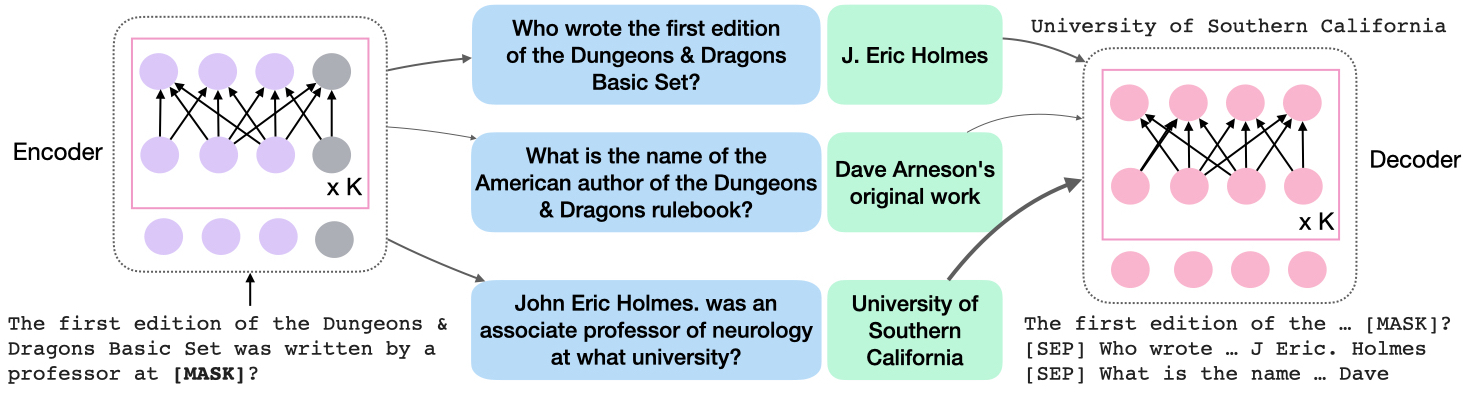}
    \caption{During pre-training, the encoder first encodes textual input and use special token representation to query the QA-memory. The retrieved QA-pairs are integrated to the decoder to generate outputs.}
    \vspace{-2ex}
    \label{fig:architecture}
\end{figure*}

However, question-retrieval QA systems have several limitations as well. 
First, there is \textit{no large-scale supervised data for question-question retrieval}.  This contrasts with the step of retrieving text given a question, where supervised data is used to build retrievers like DPR \cite{karpukhin2020dense}. To address this, RePAQ uses a latent-retrieval training process (similar to REALM \cite{lee2019latent}), in which the retriever is trained using the downstream end loss from the QA task. This requires asynchronously updating the index as training proceeds, a process that is complex and computationally expensive.  This is also a problem for domains with limited QA data: as we will show, RePAQ's performance is disappointing on smaller datasets like WebQuestions~\cite{berant2013semantic}, containing only 3K training instances.  To address this problem, we introduce a novel \textit{pre-training task for question retrieval}, which can be applied to any text-QA dataset, which great improves performance on smaller datasets.

A second problem is that RePAQ is limited to answering questions explicitly stored in the index, or paraphrases of such questions. This contrasts with QA systems that retrieve from KBs, which can typically generate complex queries that combine the atomic triples in the KB.To address this, we present an extended model that answers multi-hop questions by iteratively retrieving from a question-answer corpus, the first question-retrieval-based QA system that addresses this task.

In more detail, we propose a new QA-Memory-Augmented Transformer (\modelname) with better compositionality paired with a lower complexity training strategy. \modelname is based on a T5 encoder-decoder~\cite{raffel2020exploring} paired with an integrated key-value memory~\cite{khandelwal2019generalization,borgeaud2021improving} populated with question-answer pairs (See~\autoref{fig:architecture}). Given an input, the encoder generates a query representation scored against the QA memory and retrieves the top-K relevant QA pairs. The encoder then reprocesses the input along with the retrievals forming a QA-injected representation which is passed to the decoder to attend to and generate. 

To reduce the training (fine-tuning) sample complexity, we propose to first pre-train \modelname on a large-scale corpus to teach the model to retrieve and interpret QA pairs. We construct the pre-training corpus by leveraging existing methods for question generation, producing a very large set of potentially interesting questions from text passages ~\cite{zhou2017neural,alberti2019synthetic,lewis2021paq}. For each QA pair and the passage it was generated from, we mask the answer and train the model to fill the mask by retrieving and using an appropriate QA pair.
We show that pre-training greatly boosts the model's performance and helps the model generalize to different domains. For example, the pre-trained model can achieve a zero-shot performance of 40\% EM on NQ and TriviaQA without \emph{any} fine-tuning.

The effectiveness of this pre-training task means that we can avoid the expensive latent training procedure used by RePAQ, and instead use an efficient two-stage training pipeline. In the first stage, we use a small local in-batch memory of QA pairs to optimize the QA pair encoder.  We then freeze the encoder and construct the index for the global memory. In the second stage, we retrieve from this fixed global memory and continue to optimize the remaining parameters---including the parameters used to construct queries to the global memory---for better performance.

Lastly, we extend \modelname to build \modelnamemultihop, which iteratively retrieves from the memory to generate outputs. We demonstrate that \modelnamemultihop effectively chains multiple QA-pairs together to answer multi-hop questions in HotpotQA~\cite{yang2018hotpotqa} and Musique~\cite{trivedi2021musique}. Such compositional reasoning capability is nonexistent in RePAQ~\cite{lewis2021paq}.

In summary, we develop a new \textbf{QA augmented architecture} which extends the lines of research considering QA pairs as a representation of knowledge as well as those on memory-augmented language models. When paired with our proposed \textbf{pretraining strategy} (\autoref{sec:training}), we address many of the shortcomings of previous QA-indexing-based approaches leading to lower sample complexity training and the ability to perform \textbf{compositional reasoning} (\autoref{sec:cascade}).

\section{Related Work}
\subsection{Retriever-Reader Models}
Retrieve-and-read models have been widely studied to address knowledge-intensive tasks and achieve state-of-the-art performance on most QA tasks. These methods use two models, one to retrieve from a passage index based on BM25~\cite{robertson2009probabilistic}, and one to perform reading comprehension on the returned passages  \citep{chen-etal-2017-reading}. More recently, deep retrieval models have gained more popularity to replace traditional string-similarity retriever.

DPR~\cite{karpukhin2020dense} is a widely used supervised approach to achieve better results than BM25 on a large collection of text retrieval tasks~\cite{thakur2021beir}. Contrastive learning is used to train the deep retriever model to distinguish between annotated positive and mined negative candidates. More recently, ColBERT~\cite{khattab2020colbert} has been proposed to integrate more fine-grained late fusion between query and context to improve DPR. 
%Another line of retriever training is based on self-supervision without requiring annotated data. For example, ICT~\cite{lee2019latent}, REALM~\cite{pmlr-v119-guu20a} and  SPIDER~\cite{ram2021learning} proposed to optimize the retriever on pseudo retrieval data in either latent or supervised fashion.

Retrieval Augmented Generation (RAG)~\cite{lewis2020retrieval}, Fusion-in-Decoder (FiD)~\cite{izacard2021leveraging} and End-to-end training of Multi-Document Reader and Retriever (EmDR)~\cite{singh2021end} are proposed to read retrievals to extract or generate answers. These models require a trained retriever/reranker to obtain top-K results, which are fed to the reader to generate the answer. As discussed in~\autoref{sec:intro}, our model provides better interpretability due to atomic knowledge representation. In~\autoref{sec:main_results}, we also demonstrate that our model's inference speed is 5x faster.

\subsection{Question Generation}
The problem of question generation~\cite{zhou2017neural} has attracted attention from the community in recent years. It has been used for data augmentation~\cite{alberti2019synthetic} to improve current QA systems or to improve retrieval systems~\cite{nogueira2019document}. \citet{pan2021unsupervised} also demonstrated that by connecting generated single-hop questions, we can train zero-shot multi-hop question answering systems. Besides QA, it has also been widely used in other domains like evaluating factual consistency of summarization~\cite{eyal2019question,wang2020asking} or enhancing contextualized representation~\cite{jia2021question}. Most related to our work is PAQ~\cite{lewis2021paq}, which aims to generate and use QA pairs as retrieval units for question answering. The efficacy of this data was further verified when it was used to train DPR, yielding better domain generalization~\cite{ouguz2021domain}. 

\subsection{Memory-Augmented Language Models}
End-to-end memory-augmented language models aim to train a model to explicitly access external memory. The current work is focused on storing entities~\cite{fevry2020entities}, entity mentions~\cite{dhingra2019differentiable,sun2021reasoning,de2021mention} or knowledge triples~\cite{sun2021adaptable}. Memory attention layers are then used to influence the computation of transformer layers. These entities and fact-centric memories are naturally atomic and interpretable, and models employing them have shown competitive performance on entity-focused QA datasets like Web-Question-SP~\cite{yih2016value} and Complex\-Web\-Questions~\cite{talmor2018web}. However, these models are limited to integrating entity-centric knowledge and classifying the answer w.r.t a pre-defined entity list. For example, these models cannot handle questions with non-entity answers, e.g. number, date, noun phrases, etc, which are ubiquitous in various QA datasets like NQ~\cite{kwiatkowski2019natural}, SQuAD~\cite{rajpurkar2016squad}, or HotpotQA~\cite{yang2018hotpotqa}.

\section{Our Model: \modelname}
\begin{figure*}[!t]
    \centering
    \includegraphics[width=0.95\linewidth]{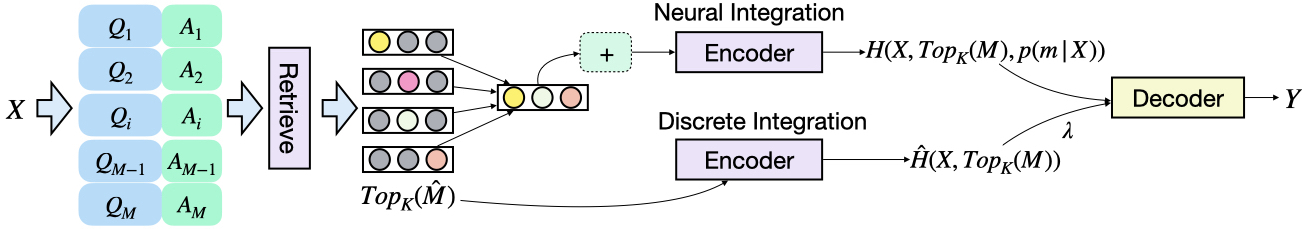}
    \caption{Architecture: upper figure shows the retrieval process with shared encoder, the lower figure shows the decoder process to leverage neural and discrete representation of memory retrieval.}
    \label{fig:full_model}
    \vspace{-2ex}
\end{figure*}
\subsection{Problem Definition}
The input to our model is a piece of text $X=x_1,\cdots,x_n$, where $X$ is either a question during fine-tuning or a paragraph in pre-training. Pre-training is formulated as a span corruption task~\cite{raffel2019exploring}: given an example in the pre-training corpus as $(X, \{Q^k, A^k\}_{k=1}^m )$, where ${A^1, \cdots, A^m}$ correspond to spans in the input $X$. We sample $k$ spans from $X$ as a cloze answer and replace all tokens within a span with a [MASK] token, and the model needs to recover all the answers. During fine-tuning, we add an artificial [MASK] in the question front, and let the model recover this as the answer. The pre-training/fine-tuning objective function is to maximize the masked language model objectives $p(Y|X) = \sum_{m_i \in M} p(Y|X, m_i) p(m_i|X)$, which marginalizes over the entire memory $M$. However, due to its intractability in a large-scale memory, we adopt an approximation to only sum over the top-K memory entries $Top_K(M)$.

We define the encoder function as $f_{\theta}$, which takes an input sequence $X$ as input to generate a sequence of vector $\mathcal{F}_{\theta}(X) \in \mathbb{R}^{n \times d}$, where $n$ is the input length and $d$ is the hidden size. The designated position of $\mathcal{F}_{\theta}(X)$ will be used as the query and memory representation, which are denoted as $f_{\theta}(X; \text{[MASK]}) \in \mathbb{R}^d$ (at [MASK] position) and memory key/value as $f_{\theta}(m_i^k; \text{[CLS]}) \in \mathbb{R}^d$ (at [CLS] position). For brevity, we leave out [MASK] and [CLS] and simply use $f_{\theta}(\cdot)$.

We also define a broadcast operator $B^n_k(x)$ to broadcast a vector into a matrix by assigning the vector $x$ to $k$-th row while filling the rest with zero, i.e.  $B^n_k(x)=[\mathbf{0}, ... x^T, ... ,\mathbf{0}]$.  

\subsection{Dense Retriever \label{sec:model_retriever}}
The memory $M$ contains separate key and value components, where the key $m_i^k$ contains a question, and the corresponding value $m_i^v$ contains the question-answer concatenation. To retrieve the top-k QA-pairs from the memory, we use our encoder $f_{\theta}$ to encode $X$ and $m_i$ separately and select the top-K entries $Top_K(M)$ based on their inner product, i.e. ${TopK}_{m_i \in M} f_{\theta}(X) \cdot f_{\theta}(m_i^k)$.

\subsection{Neural Memory Integration \label{sec:enc-dec}}
After the model retrieves the Top-K candidates, their corresponding memory values $m_i^v$ needs be leveraged into the encoder to influence the decoder outputs in a differentiable fashion. We write our objective $p(Y|X)$ as:
\begin{align*}
\small
\begin{split}
    & \sum_{m_i \in Top_K(M)} p(Y|X, m_i) p(m_i|X)  \\
   = & \sum_{m_i \in Top_K(M)} p(m_i|X) g_{\theta} (Y | \mathcal{F}_{\theta}(X) + B^n_k[f_{\theta}(m_i^v)]) \\
   \approx & g_{\theta} (Y | \sum_{m_i \in Top_K(M)} p(m_i|X)  (\mathcal{F}_{\theta}(X) + B^n_k[f_{\theta}(m_i^v)])) \\
   = & g_{\theta} (Y | \mathcal{F}_{\theta}(X) + B^n_k[\sum_{m_i \in Top_K(M)} p(m_i|X) f_{\theta}(m_i^v)]) \\
   & p(m_i|X) = \frac{e^{f_{\theta}(X) \cdot f_{\theta}(m_i^k)}}{\sum_{m_i \in TopK_{M}} e^{f_{\theta}(X) \cdot f_{\theta}(m_i^k)}}
\end{split}
\end{align*}
The probability $p(Y|X, m_i)$ is parmeterized by a decoder function $g_{\theta}$, which takes a memory-infused encoder representation $\mathcal{F}_{\theta}(X) +   B^n_k[f_{\theta}(m_i^v)]$ as input. We approximate this marginal probability by pulling weighted summation inside the decoder function $g_{\theta}$ to derive an aggregated memory-infused encoder representations $\mathcal{F}_{\theta}(X) + B^n_k[\cdots]$. The retrieval weight $p(m|X)$ is calculated as the softmax over the retrieval score over top-K items. For simplicity, $H(X, Top_K(M), p(m|X))$ is used to denote this encoder representation, thus the objective can be written as follows:
\begin{align}
\small
\begin{split}
p(Y|X) = g_{\theta} (Y | H(X, Top_K(M), p(m|X)))
\end{split}
\end{align}

As shown in the upper part of~\autoref{fig:full_model}, we first use weighted-sum over the neural representation of retrieved memory entries $f_{\theta}(m_i^v)$ and then simply add it to the encoder representation to infuse the retrieved QA-pair information. These two operations are both differentiable, which makes the it possible to train retriever latently. In essence, the retriever will increasing weights $p(m_i|X)$ on more relevant memory items instead of irrelevant ones.

%similar to prior work~\cite{fevry2020entities,sun2021adaptable}. . The advantage of this approach is that the retrieval scores $p(m|X)$ are directly leveraged into the computation graph, making the framework end-to-end differentiable. This contrasts with other prior work \cite{lee2019latent} in which the discrete tokens of retrieved documents are used.

\subsection{Neural + Discrete Memory Integration}
A disadvantage of adopting weighted-sum $\sum_i p(m_i|X) f_{\theta}(m_i^v) \in \mathbb{R}^{d}$ is that all the information from all of the top-K documents are overly compressed into a $d$-dimension vector, whereas the token retrieval representation contains more information. Therefore, we propose to add a fine-grained token-wise representation $\hat{H}(X, Top_K(M))$ to help the model access the retrieved discrete values $m_i$ directly. The representation is obtained by encoding the concatenation of the input $X$ and retrieved discrete tokens $\hat{X} = \text{Concat}[m_k;\cdots;m_1;X] \in \mathcal{R}^{(n + k|m|) \times d}$.

Such discrete memory integration greatly enriches the representation for $m_i$ and enables cross-attention between the query and retrieval, addressing the bottleneck problem. However, such discrete representation cannot propagate gradients back to the retriever. Finally, we propose to combine the neural memory $H(X, \cdot)$ and discrete memory $\hat{H}(X, \cdot)$ integration to combine their merits.
\begin{align}
\label{eq:obj}
\small
\begin{split}
    & p(Y|X) \\ 
    =& g_{\theta} (Y | H'(X,  Top_K(M), p(m|X)) + \lambda \hat{H}(X,  Top_K(M)))
\end{split}
\end{align}
where the $\lambda$ is the balancing factor to weight the two representations. We use $H'(...)=[\mathbf{0};H(...)]$ to represent the concatenation of zero-matrix $\mathbf{0} \in \mathbb{R}^{k|m|\times d}$, which has consistent dimension with $\hat{H}$. After leveraging $\hat{H}$, our model demonstrates significant improvements on the downstream tasks with 14\% on TriviaQA and 10\% on HotpotQA.

$H(X, Top_K(M), p(m|X))$ is only used to latently train the retriever, after training, we can drop it and only use the concatenated representation $\hat{H}(X, Top_K(M))$ as the encoder representation. The decoder $g_{\theta}$ will attend to $\hat{H}$ and perform a greedy search over vocabulary to generate output.

\subsection{Multi-hop Extension \label{sec:cascade}}
\begin{figure}[!h]
    \centering
    \includegraphics[width=1.0\linewidth]{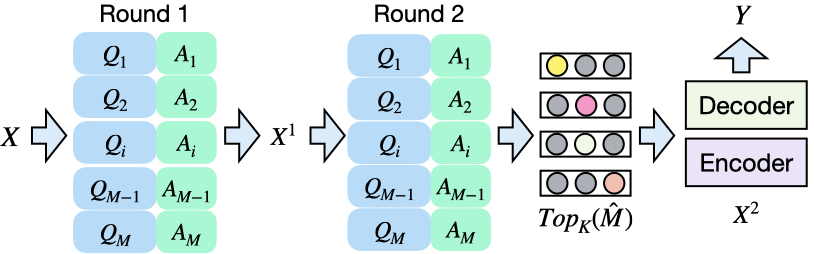}
    \caption{QAMAT+ architecture: Multi-Hop framework for question-answer memory integration.}
    \label{fig:multihop}
    \vspace{-2ex}
\end{figure}

To further extend \modelname's capability to perform compositional reasoning, we propose a cascaded architecture (depicted in~\autoref{fig:multihop}) known as \modelnamemultihop, where the model learns to perform multiple rounds of retrieval before feeding the augmented inputs to the decoder. Specifically for two-hop reasoning, we use $X$ as the query to retrieve a first-round of top-K memory values $Top_K(M;1)$ with our learned retriever $f_{\theta}$ described in \autoref{sec:model_retriever}. Next, we augment the query by concatenating the retrieved values as $X^1 = [Top_K(M;1); X]$. This new query $X^1$ is used to perform a second round of retrieval to obtain additional top-K memory values, $Top_K(M;2)$. Based on $Top_K(M;2)$, we compute the hybrid encoder representation $H(X^1, Top_K(M;2))$ and $\hat{H}(X^1, Top_K(M;2))$ to compute $p(Y|X^1;\theta)$.

\section{Training \label{sec:training}}
\subsection{Pre-training Corpus \label{sec:pretraining_corpus}}
Our QA-pairs are constructed by combining 30M deduplicated QA-pairs from PAQ~\cite{lewis2021paq}(originally 65M, we delete paraphrases to keep a subset) and 30M additional QA-pairs generated from our own pipeline. The additional QA-pairs are populated from non-overlapping passage blocks to increase the knowledge coverage over Wikipedia. Our QA generation pipeline is similar to~\cite{lewis2021paq} but trained solely on SQuAD 2.0~\cite{rajpurkar2018know} and filtered with a cheap reading comprehension model rather than FiD~\cite{izacard2021leveraging}, the details are described in the Appendix. The final statistics of our QA-memory is described in~\autoref{tab:stat}, where the total size is comparable to RePAQ.

\begin{table}[!thb]
\centering
\small
\begin{tabular}{cccc}
\toprule
Memory             & Size & \#Passages   & Training Data \\ \midrule
Dedup-PAQ                &  30M  & 10M & NaturalQuestions \\
Additional               &  30M  & 10M & SQuAD 2.0 \\
\midrule
Combined &  60M  & 20M & - \\
\bottomrule
\end{tabular}
\caption{The breakdown statistics of our QA corpus.}
\label{tab:stat}
\vspace{-2ex}
\end{table}
We denote the entire memory as $M$ and formulate the pre-training corpus as $\{X, \{Q^k, A^k\}_{k=1}^m \}$, where $X$ is the passage aligned with multiple QA-pairs $\{Q^k, A^k\}_{k=1}^m$ generated from it.

\subsection{End-to-End Training}
\begin{figure}[!tb]
    \centering
    \includegraphics[width=0.98\linewidth]{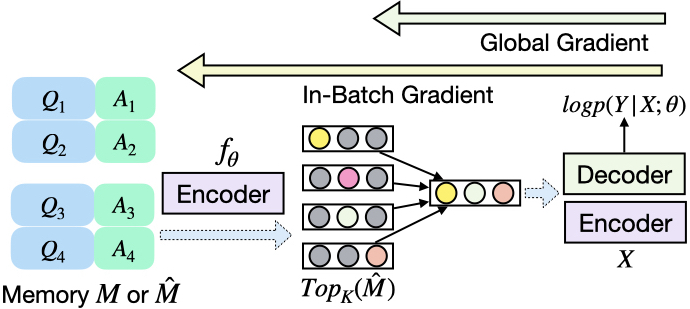}
    \caption{Two stage training procedure: in-batch training with a batch-specific memory and end-to-end gradient updates, global training with a fixed global memory and partial gradient updates.}
    \vspace{-2ex}
    \label{fig:training}
\end{figure}

During training, the retrieval process is integrated into the model's training loop. The most widely adopted approach to accomplish this is approximate nearest neighbor search (ANNS) efficiently implemented by several libraries like ScaNN~\cite{guo2020accelerating}, FAISS~\cite{johnson2019billion}, etc. These libraries require a fixed set of dense vectors to construct the index and perform a Nearest-Neighbor search using approximate algorithms. However, our memory encoder $f_{\theta}$ is continuously updated, which poses great challenges for ANNS index building. REALM~\cite{pmlr-v119-guu20a} and RePAQ~\cite{lewis2021paq} use an asynchronous index building sub-process to refresh the index every K steps, which is known to be extremely computationally expensive, especially with a large memory. To avoid such expensive computation overhead, we are inspired by TOME~\cite{de2021mention} to adopt a two-stage training as shown in~\autoref{fig:training}.

\paragraph{In-Batch Pre-training}
In the first stage, instead of using the whole memory, we propose a batch-specific memory that concatenates the positive, random negative, and hard negative entries from each instance in the batch. Assuming we have a batch size of $B$ containing examples $\{X_i, \{Q_i^k, A_i^k\}_{k=1}^{K} \}_{i=1}^{B}$. For each example there exist $K$ positive QA-pairs generated from the given context $X_i$. Additionally, we mine $K$ hard negative QA-pairs $\{\bar{Q_i^k}, \bar{A_i^k}\}_{k=1}^K$ for each input $X_i$ to increase retrieval difficulty. This hard negative mining is done with BM25~\cite{robertson2009probabilistic} similar to DPR~\cite{karpukhin2020dense}. We construct the in-batch memory by aggregating the $K \times B$ positive QA-pairs and $K \times B$ hard negative memory entries, so the in-batch memory $\hat{M}$ contains a total of $2 K \times B$ QA-pairs (roughly a few thousand). Due to the small size of the memory, we can construct the memory index very efficiently. Thus, it enables us to continuously update the memory encoder parameters $f_{\theta}$ to achieve strong QA-pair retrieval performance.

\paragraph{Global Pre-training and Fine-Tuning}
In this stage, we first freeze the memory encoder $f_{\theta}$ to generate memory-key embedding for the entire memory to build its index. We then incorporate the on-device approximate search algorithm\footnote{\small{\url{https://github.com/google-research/language/tree/master/language/mentionmemory}}} to perform the nearest-neighbor search over the memory index to retrieve the top-K QA-pairs. Formally, we propose to maximize the same objective as \autoref{eq:obj} but with stop-gradient applied to $p(m|X)$ term.
%\begin{align*}
%\small
%\begin{split}
%p(Y|X) = g_{\theta} (Y |& H^{E}(X,  Top_K(M), \mathcal{SG}(p(m|X))) \\
%                        & + \lambda \hat{H}(X,  Top_K(M)) )
%\end{split}
%\end{align*}
In this step, the model will only update the query model $f_{\theta}$ and the decoder model $g_{\theta}$. 
During fine-tuning, we follow the same recipe as the global pre-training. Instead of feeding masked passages as inputs, we use questions with pseudo [MASK] token in the front as the input.

\subsection{Multihop Extension}
For our extension model \modelnamemultihop, since the retrieval augmentation process cannot be learned latently, i.e. the gradient propagation is blocked in the concatenation step, we add additional supervision to maximize the groundtruth retrieval probability $p(m_1|X)$ for the first-round retrieval $m_1$. We add such retrieval supervision objective to the original objective $p(Y|X^1)$, where $X^1$ is the retrieval-augmented inputs as described in~\autoref{sec:cascade}.

%Formally, we optimize a combined objective function as follows:
%\begin{align*}
%\small
%\begin{split}
%   & argmax_{\theta} [log p(m_1|X;\theta) + log p(Y|X^1; \theta)]; \\
%   & where \quad p(Y|X^1; \theta)\\ 
%   = &  g_{\theta}(Y|H(X^1, Top_K(M; 2)) + \lambda \hat{H}(X^1, Top_K(M; 2)))
%\end{split}
%\end{align*}
%However, since there exists no annotated QA-pair $m_1$ in the existing datasets, we use lexical-match-based heuristics to mine the most likely ones as our silver training signal $m_1$ to optimize the first round retrieval $p(m_1|X;\theta)$. The mining algorithm is depicted in Appendix.

\section{QA Experiments}

\subsection{Implementation Details}
Our model is based on the T5-base or large architecture implemented in JAX\footnote{\small \url{https://github.com/google-research/t5x}} and pre-trained on 32 TPUs on Google Cloud\footnote{\small \url{https://cloud.google.com/tpu/}}. During in-batch training, our query and index encoder $f_{\theta}$ are shared and initialized from the T5 encoder (during global training the index encoder is fixed and the query encoder continues to be updated). Our decoder $g_{\theta}$ is similarly initialized from the T5 decoder. In total, we construct $\sim$ 60M question-answer pairs as the global memory. The memory key is the question tokenized by T5 sentencepiece model into 32 tokens, and the memory value is the answer concatenated with its question tokenized into 40 tokens. The memory is indexed by a pre-computed matrix $M^k \in \mathbb{R}^{|M| \times d}$ computed based on its keys (questions). The corresponding top-K memory values (question+answer) will be fetched. 

During in-batch pre-training, we use a large batch size of 512 and a learning rate of 1e-3, where each example contains a positive Q-A pair and 7 hard negative QA-pairs mined through BM25~\cite{robertson2009probabilistic}. The in-batch memory contains a total of 4096 entries, we set Top-k of 4 and update over all the modules. After 100K steps of in-batch pre-training, we switch to global pre-training with global memory retrieval. We decrease the batch size to 32 and enlarge Top-K to 16 for larger memory. We update only the query encoder and decoder for another 100K steps. Finally, we set K to 32 to fine-tune on downstream datasets with a decreased learning rate of 5e-4.

\subsection{Datasets}
We evaluate our framework on the three most widely used single-hop open-domain question-answering datasets and two multi-hop open-domain question-answering datasets \vspace{1ex}\\
\textbf{NQ-Open}
The NaturalQuestions~\cite{kwiatkowski2019natural} dataset consists of naturally occurring Google queries and their answers. We follow~\citet{lee2019latent} to keep questions that have a "short answer type". It consists of 79168 training examples, 8757 dev examples, and 3610 test examples.  \vspace{1ex}\\
\textbf{TriviaQA}
The TriviaQA dataset is a collection of trivia question-answer pairs that were scraped from the web~\cite{joshi2017triviaqa}. We use their unfiltered version to evaluate our model consisting of 78785 training, 8837 dev, and 113313 test examples. \vspace{1ex}\\
\textbf{WebQuestions}
The WebQuestion dataset contains questions that were sampled from Google Suggest API~\cite{berant2013semantic}. The answers are annotated from FreeBase, the training set contains 3417 examples, the dev set contains 361 examples, and the test set contains 2032 examples. \vspace{1ex}\\
\textbf{HotpotQA} The HotpotQA dataset contains questions generated by human workers by reading two passages~\cite{yang2018hotpotqa}. The questions are designed to require multiple hops and include both bridge questions and comparison questions. The training set contains a total of 90564 examples, the dev-set contains 7405 examples for evaluation. \vspace{1ex}\\
\textbf{Musique} The Musique dataset contains questions created by composing multiple questions from existing single-hop questions and was constructed to contain less bias and artifacts~\cite{trivedi2021musique}. In our experiments, we consider only the subset of 2-hop questions, resulting in a training set of 14376 examples and a dev set of 1252 examples for evaluation. While the dataset was originally designed as a distractor setting (given a question and a small number of passages, return the answer), we instead consider an open-domain setting. 

\subsection{Baselines}
We compare our model with baselines from the following categories. 1) CBQA large language models (T5 XXL), which directly outputs an answer without retrieval. 2) Entity/KG memory-augmented models that use memory attention to incorporate entity-level features into language models (Entities-as-Experts (EaE) \citep{fevry2020entities}, Fact-Injected Language Model (FilM) \citep{sun2021adaptable}, MentionMemory (TOME) \citep{de2021mention}). 3) Retrieve-and-read model, which retrieves passages to pass to a reader model which predicts the answer. 4) QA-retrieval models, which train a retriever to collect QA-pairs from a large datastore, and then rerank these QA-pairs (top 50-100) with original query with cross-attention. The highest-ranked answer is returned as the final answer.

\subsection{Single-Hop Results}
\label{sec:main_results}
Our results are summarized in~\autoref{tab:main} which reports exact-match (EM) score. \vspace{1ex}\\
\textbf{Comparison with RePAQ}
Our main comparison is with the previous best QA-retrieval-based approach "RePAQ w/ rerank (XXL ALBERT)". This model has a similar number of parameters to \modelname (Large). Without using an explicit re-ranking procedure, our model performs slightly worse on NQ but obtains significant gains on TriviaQA and WebQuestion. Especially on WebQuestion, which only contains 3K training examples, RePAQ performs significantly worse than the other datasets because it requires a high volume of examples to update the retriever from scratch. With our proposed pre-training strategy, \modelname can initialize from a much better checkpoint to decrease the sample complexity, yielding an absolute 6\% EM improvement. Additionally, without any fine-tuning, we demonstrate that our model already achieves promising results across these datasets, nearly matching the performance of "RePAQ w/o rerank"\footnote{It's worth noting that the question generation models are trained using some of these datasets' training data so this is not truly ``zero-shot'' performance.}.  \vspace{1ex}\\
\textbf{Comparison with retrieve-and-read models}
In comparison to this class of model, \modelname roughly matches the performance of RAG, though it still lags behind the SoTA model FiD. However, FiD requires reading 100 passages, i.e. 20K tokens while our best model works more efficiently by only reading top-32 QA-pairs, i.e. 1.2K tokens. To investigate the speed difference between these approaches, we compared their inference speeds using the same hardware (32 Google Cloud v3 TPUs). We found that \modelname can answer 240 Qs/sec, while FiD only answers 50 Qs/sec, a \textbf{5x inference time speedup} over FiD.

\begin{table}[!tb]
\small
\begin{tabular}{@{}llll@{}}
\toprule
Model (Test Set)         & NQ & TQA & WQ \\ \midrule
T5-3B~\cite{roberts2020much}        & 30.4             & 35.1     & 33.6         \\
T5-11B~\cite{roberts2020much}        & 32.6             & 42.3     & 37.2         \\
EaE~\cite{fevry2020entities}           & -                & 43.2     & -            \\
FILM~\cite{sun2021adaptable}          & -                & 29.1     & -            \\
TOME-2~\cite{de2021mention} & -                & 53.4     & -            \\
\midrule
DensePhrases~\cite{lee2021learning}   & 40.9 & 50.7 & - \\
REALM~\cite{pmlr-v119-guu20a}         & 40.4             & 55.8        & 40.7         \\
DPR~\cite{karpukhin2020dense}           & 41.5             & 57.9     & 42.4         \\
RAG-Seq~\cite{lewis2020retrieval}       & 44.5             & 56.8     & 45.2         \\
FiD~\cite{izacard2021leveraging}           & 48.2             & 65.0     & -            \\
\midrule
RePAQ~\cite{lewis2021paq} & 41.2  &  38.8   &  29.4$\dagger$ \\
RePAQ+Rerank~\cite{lewis2021paq} & 47.6  &  50.7   &  37.6$\dagger$ \\
\midrule
\modelname Zero-Shot (Base)   & 37.9             & 34.1     & 25.9         \\
\modelname Zero-Shot (Large)  & 39.8             & 40.0       & 25.1        \\ 
\modelname Fine-tuned (Base)   & 44.5             & 53.2     & 43.0         \\
\modelname Fine-tuned (Large)  & 45.5             & 54.8     & 43.6         \\
\bottomrule
\end{tabular}
\caption{The main experimental results on single-hop question answering datasets (NQ=NaturalQuestions, TQA=TriviaQA, WQ=WebQuestions), $\dagger$ means Best-effort replication using our own implementation.}
\label{tab:main}
\end{table}

\subsection{Multi-hop Results}
Since the document corpora source of HotpotQA and Musique are different from single-hop QA datasets, we adopt question generation model trained on SQuAD 2.0~\cite{rajpurkar2018know} to generate questions for these two datasets. To create the document corpora, we gather all of the provided positive and negative documents, obtaining 500K passages for HotpotQA and 85K passages for Musique. We then use the trained generation models to populate 3M QA pairs for HotpotQA and 500K QA pairs for Musique. These QA pairs are then used as the memory source for \modelnamemultihop, simulating a (slightly smaller) open-domain setup. When training \modelnamemultihop on Musique, we initialize from HotpotQA's in-Batch pre-trained checkpoint, which can bring 5-7\% F1 improvement. 

\begin{table}[!tb]
\small
\begin{tabular}{lcc}
\toprule
Model (Dev Set F1 Score)         & HPQ & MusQ  \\ \midrule
T5-3B~\cite{roberts2020much}        &  27.8            &    7.5       \\
T5-11B~\cite{roberts2020much}        &  30.2            &   9.0  \\
\midrule
MDR+T5-Decoder~\cite{xiong2020answering}  & \textbf{62.6} & 26.8  \\
RePAQ~\cite{lewis2021paq}$\dagger$   & 47.8  &  18.6 \\
\midrule
\modelname       &    42.0     &  16.7      \\
\modelnamemultihop &  57.6     &  \textbf{29.8}       \\
\bottomrule
\end{tabular}
\caption{The main experimental results on MultiHop QA datasets with \modelname and \modelnamemultihop, $\dagger$ means Best-effort replication using our own implementation.}
\label{tab:multihop}
\end{table}

In~\autoref{tab:multihop}, we show that \modelnamemultihop achieves promising results on both multi-hop datasets, outperforming T5-CBQA and RePAQ by a large margin. Additionally, \modelnamemultihop performs considerably better than the single-hop \modelname, demonstrating the effectiveness of performing multi-round retrieval. Though \modelnamemultihop still lags behind the document-based model (MDR+T5 Decoder) on HotpotQA, it surpasses it on the more challenging Musique dataset.  These encouraging results suggest the potential for \modelnamemultihop to perform compositional reasoning over multiple QA-pairs, which greatly increases the coverage of QA datastore to cover more composite factual information. 

\subsection{Ablation Studies \label{sec:ablations}}
\begin{table}[!tb]
\small
\centering
\begin{tabular}{lllll}
\toprule
Top-K            & 1   & 10    & 20   &   30   \\
\midrule
NQ-Recall@K       & 0.41  & 0.58  & 0.62 & 0.64 \\
TriviaQA-Recll@K & 0.46 & 0.66  & 0.70  & 0.72 \\
\midrule
NQ-EM@K        & 0.39 & 0.42 & 0.44 & 0.44 \\
TriviaQA-EM@K  & 0.45 & 0.51  & 0.53 & 0.53 \\
\bottomrule
\end{tabular}
\caption{The retrieval recall and EM score of different retrieval numbers on test sets.}
\label{tab:retrieval}
\end{table}

\begin{table}[!tb]
\centering
\small
\begin{tabular}{@{}llll@{}}
\toprule
Pre-training Stages        & NQ & TQA & WQ \\ 
\midrule
Only In-Batch     &  42.1            &  48.2    &   39.7      \\
Only Global       & 26.0             & 28.9     &  26.1      \\
In-Batch $\rightarrow$ Global     & 44.5             & 53.2     & 43.0        \\
\bottomrule
\end{tabular}
\caption{Downstream EM performance of models when pre-trained using in-batch, global, or both stages.}
\label{tab:ablation_v2}
\vspace{-3ex}
\end{table}

\textbf{Number of Retrievals}
To understand the properties of our model better, we first investigate the impact of the number of retrievals, $K$, on the model's performance. We gradually increase the $K$ to collect the recall and final QA performance. The results are shown in~\autoref{tab:retrieval}. We observe that even though retrieval recall continues to increase beyond $K > 20$, the EM score saturates much earlier. Future research could improve performance further by developing decoders to more accurately exploit these larger retrievals sets. \vspace{1ex} \\
\noindent \textbf{Importance of Two-Stage Pre-training}
We next analyze the importance of the two-stage pre-training from \autoref{sec:training} by removing either the in-batch or global stage. From our results shown in~\autoref{tab:ablation_v2}, we can see that using in-batch pre-training alone leads to a degradation in performance when compared to the two-stage approach. This is likely because the model is never exposed to the full set of hard negatives which will be encountered when performing retrieval over the global memory. On the other hand, if we directly pre-train the global-memory model without any in-batch initialization, the retriever performance is nearly random and the decoder consequently learns to ignore the retrieval and simply memorize question-answer pairs.

\section{Conclusion}
In this paper, we propose a more accurate and efficient architecture to utilize QA-pairs as representation units of knowledge. Our proposed model \modelname outperforms RePAQ significantly, while leveraging our less expensive training procedure. Furthermore, we show how a QA-backed model can perform compositional reasoning and address more complex queries. In the future, we hope to further close the gap with state-of-the-art document-based retrieve-and-read models and extend this approach to a broader set of tasks.

\section*{Limitations}
Our approach has several limitations: 1) we use generated question-answer pairs as a knowledge base, which are extracted from web documents. In order to maintain high quality and faithfulness, the question generation pipeline needs to be well trained with a sufficient amount of clean data. Such conditions might not hold for other domains outside of Wikipedia like biomedical text, thus the general QA-as-Knowledge-Base concept could require additional innovations to extend to other areas. 2) Our latent retrieval learning requires quasi paired data to learn the alignment between the query and memory. This is hard to satisfy in some domains with noisier data or only a very weak alignment between a query and the memory. 3) Our model requires mined intermediate retrieval signals to train \modelnamemultihop, which currently relies on lexical-overlap-based heuristics. In other cases, this may not be sufficient and instead might require a more principled design to mine better intermediate supervision. 

\section*{Ethical Statement}
Our work encourages the model to ground on the existing knowledge populated from large textual collections. We believe it is a reasonable towards building more trustworthy and more robust machine learning models. Having better attributions to knowledge source could help humans better understand the model's rationale for decision making. However, we do admit that the question generation models used to populate the QA knowledge base could potentially exacerbate the biases already present in the original Wikipedia data. We will keep working on this direction to minimize its potential negative impacts.       

\bibliography{custom}

\begin{thebibliography}{45}
\expandafter\ifx\csname natexlab\endcsname\relax\def\natexlab#1{#1}\fi

\bibitem[{Alberti et~al.(2019)Alberti, Andor, Pitler, Devlin, and
  Collins}]{alberti2019synthetic}
Chris Alberti, Daniel Andor, Emily Pitler, Jacob Devlin, and Michael Collins.
  2019.
\newblock Synthetic qa corpora generation with roundtrip consistency.
\newblock In \emph{Proceedings of the 57th Annual Meeting of the Association
  for Computational Linguistics}, pages 6168--6173.

\bibitem[{Berant et~al.(2013)Berant, Chou, Frostig, and
  Liang}]{berant2013semantic}
Jonathan Berant, Andrew Chou, Roy Frostig, and Percy Liang. 2013.
\newblock Semantic parsing on freebase from question-answer pairs.
\newblock In \emph{Proceedings of the 2013 conference on empirical methods in
  natural language processing}, pages 1533--1544.

\bibitem[{Borgeaud et~al.(2021)Borgeaud, Mensch, Hoffmann, Cai, Rutherford,
  Millican, Driessche, Lespiau, Damoc, Clark et~al.}]{borgeaud2021improving}
Sebastian Borgeaud, Arthur Mensch, Jordan Hoffmann, Trevor Cai, Eliza
  Rutherford, Katie Millican, George van~den Driessche, Jean-Baptiste Lespiau,
  Bogdan Damoc, Aidan Clark, et~al. 2021.
\newblock Improving language models by retrieving from trillions of tokens.
\newblock \emph{arXiv preprint arXiv:2112.04426}.

\bibitem[{Chen et~al.(2017{\natexlab{a}})Chen, Fisch, Weston, and
  Bordes}]{chen2017reading}
Danqi Chen, Adam Fisch, Jason Weston, and Antoine Bordes. 2017{\natexlab{a}}.
\newblock Reading wikipedia to answer open-domain questions.
\newblock In \emph{Proceedings of the 55th Annual Meeting of the Association
  for Computational Linguistics (Volume 1: Long Papers)}, pages 1870--1879.

\bibitem[{Chen et~al.(2017{\natexlab{b}})Chen, Fisch, Weston, and
  Bordes}]{chen-etal-2017-reading}
Danqi Chen, Adam Fisch, Jason Weston, and Antoine Bordes. 2017{\natexlab{b}}.
\newblock \href {https://doi.org/10.18653/v1/P17-1171} {Reading {W}ikipedia to
  answer open-domain questions}.
\newblock In \emph{Proceedings of the 55th Annual Meeting of the Association
  for Computational Linguistics (Volume 1: Long Papers)}, pages 1870--1879,
  Vancouver, Canada. Association for Computational Linguistics.

\bibitem[{de~Jong et~al.(2022)de~Jong, Zemlyanskiy, FitzGerald, Sha, and
  Cohen}]{de2021mention}
Michiel de~Jong, Yury Zemlyanskiy, Nicholas FitzGerald, Fei Sha, and William
  Cohen. 2022.
\newblock Mention memory: incorporating textual knowledge into transformers
  through entity mention attention.
\newblock \emph{International Conference on Learning Representations}.

\bibitem[{Dhingra et~al.(2019)Dhingra, Zaheer, Balachandran, Neubig,
  Salakhutdinov, and Cohen}]{dhingra2019differentiable}
Bhuwan Dhingra, Manzil Zaheer, Vidhisha Balachandran, Graham Neubig, Ruslan
  Salakhutdinov, and William~W Cohen. 2019.
\newblock Differentiable reasoning over a virtual knowledge base.
\newblock In \emph{International Conference on Learning Representations}.

\bibitem[{Eyal et~al.(2019)Eyal, Baumel, and Elhadad}]{eyal2019question}
Matan Eyal, Tal Baumel, and Michael Elhadad. 2019.
\newblock Question answering as an automatic evaluation metric for news article
  summarization.
\newblock In \emph{Proceedings of NAACL-HLT}, pages 3938--3948.

\bibitem[{F{\'e}vry et~al.(2020)F{\'e}vry, Soares, Fitzgerald, Choi, and
  Kwiatkowski}]{fevry2020entities}
Thibault F{\'e}vry, Livio~Baldini Soares, Nicholas Fitzgerald, Eunsol Choi, and
  Tom Kwiatkowski. 2020.
\newblock Entities as experts: Sparse memory access with entity supervision.
\newblock In \emph{Proceedings of the 2020 Conference on Empirical Methods in
  Natural Language Processing (EMNLP)}, pages 4937--4951.

\bibitem[{Guo et~al.(2020)Guo, Sun, Lindgren, Geng, Simcha, Chern, and
  Kumar}]{guo2020accelerating}
Ruiqi Guo, Philip Sun, Erik Lindgren, Quan Geng, David Simcha, Felix Chern, and
  Sanjiv Kumar. 2020.
\newblock Accelerating large-scale inference with anisotropic vector
  quantization.
\newblock In \emph{International Conference on Machine Learning}, pages
  3887--3896. PMLR.

\bibitem[{Guu et~al.(2020)Guu, Lee, Tung, Pasupat, and
  Chang}]{pmlr-v119-guu20a}
Kelvin Guu, Kenton Lee, Zora Tung, Panupong Pasupat, and Mingwei Chang. 2020.
\newblock \href {https://proceedings.mlr.press/v119/guu20a.html} {Retrieval
  augmented language model pre-training}.
\newblock In \emph{Proceedings of the 37th International Conference on Machine
  Learning}, volume 119 of \emph{Proceedings of Machine Learning Research},
  pages 3929--3938. PMLR.

\bibitem[{Izacard and Grave(2021)}]{izacard2021leveraging}
Gautier Izacard and {\'E}douard Grave. 2021.
\newblock Leveraging passage retrieval with generative models for open domain
  question answering.
\newblock In \emph{Proceedings of the 16th Conference of the European Chapter
  of the Association for Computational Linguistics: Main Volume}, pages
  874--880.

\bibitem[{Jia et~al.(2021)Jia, Lewis, and Zettlemoyer}]{jia2021question}
Robin Jia, Mike Lewis, and Luke Zettlemoyer. 2021.
\newblock Question answering infused pre-training of general-purpose
  contextualized representations.
\newblock \emph{arXiv preprint arXiv:2106.08190}.

\bibitem[{Johnson et~al.(2019)Johnson, Douze, and
  J{\'e}gou}]{johnson2019billion}
Jeff Johnson, Matthijs Douze, and Herv{\'e} J{\'e}gou. 2019.
\newblock Billion-scale similarity search with {GPUs}.
\newblock \emph{IEEE Transactions on Big Data}, 7(3):535--547.

\bibitem[{Joshi et~al.(2017)Joshi, Choi, Weld, and
  Zettlemoyer}]{joshi2017triviaqa}
Mandar Joshi, Eunsol Choi, Daniel~S Weld, and Luke Zettlemoyer. 2017.
\newblock Triviaqa: A large scale distantly supervised challenge dataset for
  reading comprehension.
\newblock In \emph{Proceedings of the 55th Annual Meeting of the Association
  for Computational Linguistics (Volume 1: Long Papers)}, pages 1601--1611.

\bibitem[{Karpukhin et~al.(2020)Karpukhin, Oguz, Min, Lewis, Wu, Edunov, Chen,
  and Yih}]{karpukhin2020dense}
Vladimir Karpukhin, Barlas Oguz, Sewon Min, Patrick Lewis, Ledell Wu, Sergey
  Edunov, Danqi Chen, and Wen-tau Yih. 2020.
\newblock Dense passage retrieval for open-domain question answering.
\newblock In \emph{Proceedings of the 2020 Conference on Empirical Methods in
  Natural Language Processing (EMNLP)}, pages 6769--6781.

\bibitem[{Khandelwal et~al.(2019)Khandelwal, Levy, Jurafsky, Zettlemoyer, and
  Lewis}]{khandelwal2019generalization}
Urvashi Khandelwal, Omer Levy, Dan Jurafsky, Luke Zettlemoyer, and Mike Lewis.
  2019.
\newblock Generalization through memorization: Nearest neighbor language
  models.
\newblock In \emph{International Conference on Learning Representations}.

\bibitem[{Khattab and Zaharia(2020)}]{khattab2020colbert}
Omar Khattab and Matei Zaharia. 2020.
\newblock Colbert: Efficient and effective passage search via contextualized
  late interaction over bert.
\newblock In \emph{Proceedings of the 43rd International ACM SIGIR conference
  on research and development in Information Retrieval}, pages 39--48.

\bibitem[{Kwiatkowski et~al.(2019)Kwiatkowski, Palomaki, Redfield, Collins,
  Parikh, Alberti, Epstein, Polosukhin, Devlin, Lee
  et~al.}]{kwiatkowski2019natural}
Tom Kwiatkowski, Jennimaria Palomaki, Olivia Redfield, Michael Collins, Ankur
  Parikh, Chris Alberti, Danielle Epstein, Illia Polosukhin, Jacob Devlin,
  Kenton Lee, et~al. 2019.
\newblock Natural questions: a benchmark for question answering research.
\newblock \emph{Transactions of the Association for Computational Linguistics},
  7:453--466.

\bibitem[{Lee et~al.(2021)Lee, Sung, Kang, and Chen}]{lee2021learning}
Jinhyuk Lee, Mujeen Sung, Jaewoo Kang, and Danqi Chen. 2021.
\newblock Learning dense representations of phrases at scale.
\newblock In \emph{Proceedings of the 59th Annual Meeting of the Association
  for Computational Linguistics and the 11th International Joint Conference on
  Natural Language Processing (Volume 1: Long Papers)}, pages 6634--6647.

\bibitem[{Lee et~al.(2019)Lee, Chang, and Toutanova}]{lee2019latent}
Kenton Lee, Ming-Wei Chang, and Kristina Toutanova. 2019.
\newblock Latent retrieval for weakly supervised open domain question
  answering.
\newblock In \emph{Proceedings of the 57th Annual Meeting of the Association
  for Computational Linguistics}, pages 6086--6096.

\bibitem[{Lewis et~al.(2020)Lewis, Perez, Piktus, Petroni, Karpukhin, Goyal,
  K{\"u}ttler, Lewis, Yih, Rockt{\"a}schel et~al.}]{lewis2020retrieval}
Patrick Lewis, Ethan Perez, Aleksandra Piktus, Fabio Petroni, Vladimir
  Karpukhin, Naman Goyal, Heinrich K{\"u}ttler, Mike Lewis, Wen-tau Yih, Tim
  Rockt{\"a}schel, et~al. 2020.
\newblock Retrieval-augmented generation for knowledge-intensive nlp tasks.
\newblock \emph{Advances in Neural Information Processing Systems},
  33:9459--9474.

\bibitem[{Lewis et~al.(2021)Lewis, Wu, Liu, Minervini, K{\"u}ttler, Piktus,
  Stenetorp, and Riedel}]{lewis2021paq}
Patrick Lewis, Yuxiang Wu, Linqing Liu, Pasquale Minervini, Heinrich
  K{\"u}ttler, Aleksandra Piktus, Pontus Stenetorp, and Sebastian Riedel. 2021.
\newblock Paq: 65 million probably-asked questions and what you can do with
  them.
\newblock \emph{Transactions of the Association for Computational Linguistics},
  9:1098--1115.

\bibitem[{Min et~al.(2021)Min, Boyd-Graber, Alberti, Chen, Choi, Collins, Guu,
  Hajishirzi, Lee, Palomaki et~al.}]{min2021neurips}
Sewon Min, Jordan Boyd-Graber, Chris Alberti, Danqi Chen, Eunsol Choi, Michael
  Collins, Kelvin Guu, Hannaneh Hajishirzi, Kenton Lee, Jennimaria Palomaki,
  et~al. 2021.
\newblock Neurips 2020 efficientqa competition: Systems, analyses and lessons
  learned.
\newblock In \emph{NeurIPS 2020 Competition and Demonstration Track}, pages
  86--111. PMLR.

\bibitem[{Nogueira et~al.(2019)Nogueira, Yang, Lin, and
  Cho}]{nogueira2019document}
Rodrigo Nogueira, Wei Yang, Jimmy Lin, and Kyunghyun Cho. 2019.
\newblock Document expansion by query prediction.
\newblock \emph{arXiv preprint arXiv:1904.08375}.

\bibitem[{O{\u{g}}uz et~al.(2021)O{\u{g}}uz, Lakhotia, Gupta, Lewis, Karpukhin,
  Piktus, Chen, Riedel, Yih, Gupta et~al.}]{ouguz2021domain}
Barlas O{\u{g}}uz, Kushal Lakhotia, Anchit Gupta, Patrick Lewis, Vladimir
  Karpukhin, Aleksandra Piktus, Xilun Chen, Sebastian Riedel, Wen-tau Yih,
  Sonal Gupta, et~al. 2021.
\newblock Domain-matched pre-training tasks for dense retrieval.
\newblock \emph{arXiv preprint arXiv:2107.13602}.

\bibitem[{Pan et~al.(2021)Pan, Chen, Xiong, Kan, and
  Wang}]{pan2021unsupervised}
Liangming Pan, Wenhu Chen, Wenhan Xiong, Min-Yen Kan, and William~Yang Wang.
  2021.
\newblock Unsupervised multi-hop question answering by question generation.
\newblock In \emph{Proceedings of the 2021 Conference of the North American
  Chapter of the Association for Computational Linguistics: Human Language
  Technologies}, pages 5866--5880.

\bibitem[{Raffel et~al.(2019)Raffel, Shazeer, Roberts, Lee, Narang, Matena,
  Zhou, Li, and Liu}]{raffel2019exploring}
Colin Raffel, Noam Shazeer, Adam Roberts, Katherine Lee, Sharan Narang, Michael
  Matena, Yanqi Zhou, Wei Li, and Peter~J Liu. 2019.
\newblock Exploring the limits of transfer learning with a unified text-to-text
  transformer.
\newblock \emph{arXiv preprint arXiv:1910.10683}.

\bibitem[{Raffel et~al.(2020)Raffel, Shazeer, Roberts, Lee, Narang, Matena,
  Zhou, Li, and Liu}]{raffel2020exploring}
Colin Raffel, Noam Shazeer, Adam Roberts, Katherine Lee, Sharan Narang, Michael
  Matena, Yanqi Zhou, Wei Li, and Peter~J Liu. 2020.
\newblock Exploring the limits of transfer learning with a unified text-to-text
  transformer.
\newblock \emph{Journal of Machine Learning Research}, 21:1--67.

\bibitem[{Rajpurkar et~al.(2018)Rajpurkar, Jia, and Liang}]{rajpurkar2018know}
Pranav Rajpurkar, Robin Jia, and Percy Liang. 2018.
\newblock Know what you don’t know: Unanswerable questions for squad.
\newblock In \emph{Proceedings of the 56th Annual Meeting of the Association
  for Computational Linguistics (Volume 2: Short Papers)}, pages 784--789.

\bibitem[{Rajpurkar et~al.(2016)Rajpurkar, Zhang, Lopyrev, and
  Liang}]{rajpurkar2016squad}
Pranav Rajpurkar, Jian Zhang, Konstantin Lopyrev, and Percy Liang. 2016.
\newblock Squad: 100,000+ questions for machine comprehension of text.
\newblock In \emph{Proceedings of the 2016 Conference on Empirical Methods in
  Natural Language Processing}, pages 2383--2392.

\bibitem[{Roberts et~al.(2020)Roberts, Raffel, and Shazeer}]{roberts2020much}
Adam Roberts, Colin Raffel, and Noam Shazeer. 2020.
\newblock How much knowledge can you pack into the parameters of a language
  model?
\newblock In \emph{Proceedings of the 2020 Conference on Empirical Methods in
  Natural Language Processing (EMNLP)}, pages 5418--5426.

\bibitem[{Robertson and Zaragoza(2009)}]{robertson2009probabilistic}
Stephen Robertson and Hugo Zaragoza. 2009.
\newblock \emph{The probabilistic relevance framework: BM25 and beyond}.
\newblock Now Publishers Inc.

\bibitem[{Singh et~al.(2021)Singh, Reddy, Hamilton, Dyer, and
  Yogatama}]{singh2021end}
Devendra Singh, Siva Reddy, Will Hamilton, Chris Dyer, and Dani Yogatama. 2021.
\newblock End-to-end training of multi-document reader and retriever for
  open-domain question answering.
\newblock \emph{Advances in Neural Information Processing Systems}, 34.

\bibitem[{Sun et~al.(2021)Sun, Verga, Dhingra, Salakhutdinov, and
  Cohen}]{sun2021reasoning}
Haitian Sun, Pat Verga, Bhuwan Dhingra, Ruslan Salakhutdinov, and William~W
  Cohen. 2021.
\newblock Reasoning over virtual knowledge bases with open predicate relations.
\newblock In \emph{International Conference on Machine Learning}, pages
  9966--9977. PMLR.

\bibitem[{Talmor and Berant(2018)}]{talmor2018web}
Alon Talmor and Jonathan Berant. 2018.
\newblock The web as a knowledge-base for answering complex questions.
\newblock In \emph{Proceedings of the 2018 Conference of the North American
  Chapter of the Association for Computational Linguistics: Human Language
  Technologies, Volume 1 (Long Papers)}, pages 641--651.

\bibitem[{Thakur et~al.(2021)Thakur, Reimers, R{\"u}ckl{\'e}, Srivastava, and
  Gurevych}]{thakur2021beir}
Nandan Thakur, Nils Reimers, Andreas R{\"u}ckl{\'e}, Abhishek Srivastava, and
  Iryna Gurevych. 2021.
\newblock Beir: A heterogeneous benchmark for zero-shot evaluation of
  information retrieval models.
\newblock In \emph{Thirty-fifth Conference on Neural Information Processing
  Systems Datasets and Benchmarks Track (Round 2)}.

\bibitem[{Trivedi et~al.(2021)Trivedi, Balasubramanian, Khot, and
  Sabharwal}]{trivedi2021musique}
Harsh Trivedi, Niranjan Balasubramanian, Tushar Khot, and Ashish Sabharwal.
  2021.
\newblock Musique: Multi-hop questions via single-hop question composition.
\newblock \emph{arXiv preprint arXiv:2108.00573}.

\bibitem[{Verga et~al.(2021)Verga, Sun, Soares, and Cohen}]{sun2021adaptable}
Pat Verga, Haitian Sun, Livio~Baldini Soares, and William~Weston Cohen. 2021.
\newblock Adaptable and interpretable neural memory over symbolic knowledge.
\newblock In \emph{Proceedings of NAACL-HLT}, pages 3678--3691.

\bibitem[{Wang et~al.(2020)Wang, Cho, and Lewis}]{wang2020asking}
Alex Wang, Kyunghyun Cho, and Mike Lewis. 2020.
\newblock Asking and answering questions to evaluate the factual consistency of
  summaries.
\newblock In \emph{Proceedings of the 58th Annual Meeting of the Association
  for Computational Linguistics}, pages 5008--5020.

\bibitem[{Xiao et~al.(2021)Xiao, Wang, Dernoncourt, Bui, Sun, and
  Han}]{xiao2021open}
Jinfeng Xiao, Lidan Wang, Franck Dernoncourt, Trung Bui, Tong Sun, and Jiawei
  Han. 2021.
\newblock Open-domain question answering with pre-constructed question spaces.
\newblock In \emph{Proceedings of the 2021 Conference of the North American
  Chapter of the Association for Computational Linguistics: Student Research
  Workshop}, pages 61--67.

\bibitem[{Xiong et~al.(2020)Xiong, Li, Iyer, Du, Lewis, Wang, Mehdad, Yih,
  Riedel, Kiela et~al.}]{xiong2020answering}
Wenhan Xiong, Xiang Li, Srini Iyer, Jingfei Du, Patrick Lewis, William~Yang
  Wang, Yashar Mehdad, Scott Yih, Sebastian Riedel, Douwe Kiela, et~al. 2020.
\newblock Answering complex open-domain questions with multi-hop dense
  retrieval.
\newblock In \emph{International Conference on Learning Representations}.

\bibitem[{Yang et~al.(2018)Yang, Qi, Zhang, Bengio, Cohen, Salakhutdinov, and
  Manning}]{yang2018hotpotqa}
Zhilin Yang, Peng Qi, Saizheng Zhang, Yoshua Bengio, William Cohen, Ruslan
  Salakhutdinov, and Christopher~D Manning. 2018.
\newblock Hotpotqa: A dataset for diverse, explainable multi-hop question
  answering.
\newblock In \emph{Proceedings of the 2018 Conference on Empirical Methods in
  Natural Language Processing}, pages 2369--2380.

\bibitem[{Yih et~al.(2016)Yih, Richardson, Meek, Chang, and Suh}]{yih2016value}
Wen-tau Yih, Matthew Richardson, Christopher Meek, Ming-Wei Chang, and Jina
  Suh. 2016.
\newblock The value of semantic parse labeling for knowledge base question
  answering.
\newblock In \emph{Proceedings of the 54th Annual Meeting of the Association
  for Computational Linguistics (Volume 2: Short Papers)}, pages 201--206.

\bibitem[{Zhou et~al.(2017)Zhou, Yang, Wei, Tan, Bao, and
  Zhou}]{zhou2017neural}
Qingyu Zhou, Nan Yang, Furu Wei, Chuanqi Tan, Hangbo Bao, and Ming Zhou. 2017.
\newblock Neural question generation from text: A preliminary study.
\newblock In \emph{National CCF Conference on Natural Language Processing and
  Chinese Computing}, pages 662--671. Springer.

\end{thebibliography}
\bibliographystyle{acl_natbib}

\clearpage
\appendix
\label{sec:appendix}
\section{Question Answer Pairs as Knowledge Base}
\begin{figure}[!thb]
    \centering
    \includegraphics[width=0.9\linewidth]{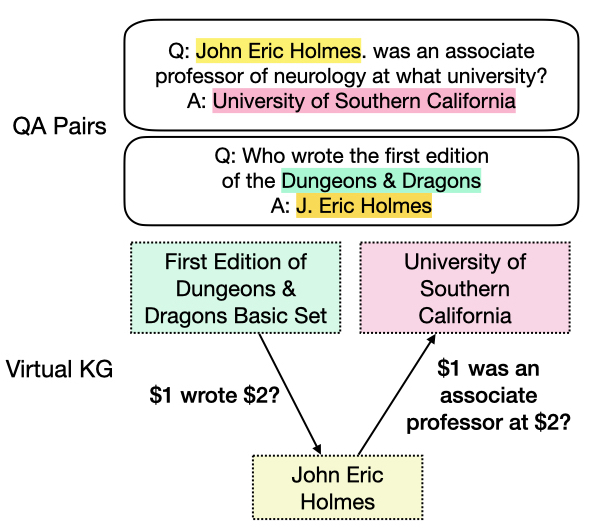}
    \caption{QA pairs can be seen as virtual knowledge base, where the question can represent complex relations connecting subject and answer.}
    \label{fig:paq}
\end{figure}

We can see QA-pairs as a virtual knowledge graph, where the question template defines the relation, the topic entity in the question defines the head entity node, and the answer denotes the tail entity. A typical example is given in~\autoref{fig:paq}, such compositionality makes the QA-pair more controllable and easy to reason over than documents. 

\section{Question Generation}
Here, we use existing SQuAD datasets' <Q, A, Document> triples~\cite{rajpurkar2016squad} to train answer extraction, question generation model. 

\paragraph{Answer Extraction}
Specifically, our answer extraction model takes a document as the input and trains an encoder-decoder model to generate a potential answer. We use beam search over the trained model to find the highest-likely answers in the given document. In our experiment, the answer extraction model is trained with the SQuAD dataset, where the document is given as the input, and the answer spans are the prediction targets. 

\paragraph{Question Generation}
For the question generation model, we take the SQuAD dataset and use document + extracted answer as the input to generate questions as the outputs. This step is also accomplished by an encoder-decoder model.  which is mainly purposed for reading comprehension problems, where the annotated questions are highly correlated with the document containing very few hallucinations. However, the questions in SQuAD~\cite{rajpurkar2016squad} could be contextualized or ambiguous, which could lead to ambiguity problems to hurt the retrieval performance. Therefore, we add question filtering to select the most accurate QA pairs.

\paragraph{Question Filtering}
For the question filtering model, we take the document + generated question to generate an answer. We compare the predicted answer vs. the original answer to see if they match each other. If not, the QA-pair will be filtered based on such inconsistency. We use a reading comprehension model trained with SQuAD to predict the answer. The predicted answer based on the document will match with the original QA-pair to decide its consistency. Such an option runs much faster, providing much higher recall but lower precision compared to the open-domain FiD filtering used in~\cite{lewis2021paq}.

\begin{figure}[!thb]
    \centering
    \includegraphics[width=1.0\linewidth]{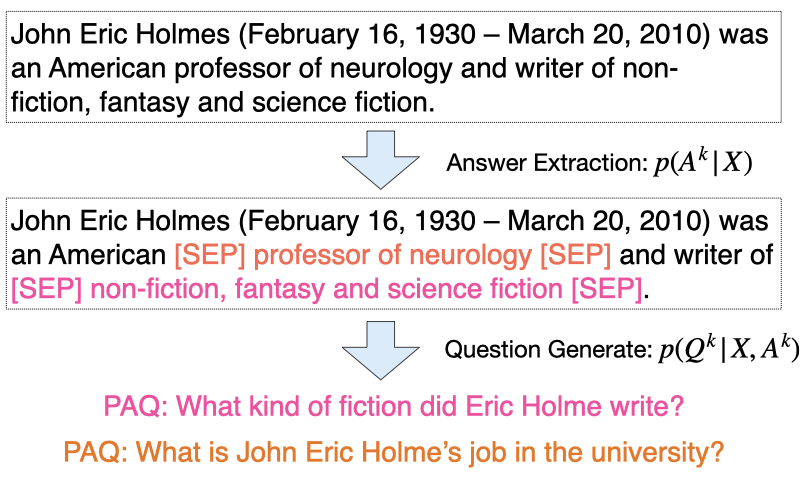}
    \caption{Question generation pipeline: Answers are extracted from passages and then questions are generated conditioned on that contextualized answer. This procedure is used to generate both our model's QA memory and our pre-training data.}
    \label{fig:qgen}
    \vspace{-3ex}
\end{figure}

We visualize our question generation pipeline mentioned above in~\autoref{fig:qgen}.

\section{Ablation Study}
We experiment with two variants of memory to see their performance difference. 
\subsection{PAQ memory}
The first version is the standard PAQ corpus~\cite{lewis2021paq} containing 65M QA pairs, where these QA-pairs are generated by models trained on NQ~\cite{kwiatkowski2019natural} and filtered through FiD model~\cite{izacard2021leveraging} also trained on NQ~\cite{kwiatkowski2019natural}. This memory is highly precise due to ODQA-filtering process, however, it only covers information from 9M out of the 20M passage blocks used in DPR~\cite{karpukhin2020dense}. 

Our memory contains 30M PAQ corpus being de-duplicated, i.e. only one question corresponds to an answer span. We generate 30M additional QA-pairs based on the left-out 10M documents from PAQ~\cite{lewis2021paq} and add these complementary QA-pairs to form our 60M memory to increase the coverage. However, since our filtering procedure is based on reading comprehension, the precision of QA-pairs is lower than the original PAQ memory.

\begin{table}[!thb]
\centering
\small
\begin{tabular}{lccc}
\toprule
Memory  & NQ             & TriviaQA     & WebQuestions         \\ 
\midrule
PAQ 65M       & 44.7             & 48.0     & 39.4         \\ 
Ours 60M   & 44.5             & 53.2     & 43.0         \\
\bottomrule
\end{tabular}
\caption{Impact of different memory over the downstream QA dataset performance.}
\label{tab:memory}
\end{table}

As can be seen, from~\autoref{tab:memory}, using the most precise but low-coverage PAQ memory from PAQ~\cite{lewis2021paq} yields the worse results on TriviaQA and WebQuestions. After adding an additional 30M PAQs to the memory generated by our pipeline, we are able to achieve 4-5\% improvements on these two datasets while still maintaining NQ's performance.

\subsection{Size of Pre-training Corpus }
Next, we investigate the impact of the size of the pre-training corpus. As a baseline, we repurpose the aligned query-passage corpus used to train DPR~\cite{karpukhin2020dense} which we adapt to our setting by simply reversing the pairs (120K passage -> question retrieval). Additionally, we vary the size of generated pre-training corpus (from 1M to 20M instances) to see its impact on the model's final downstream performance. From~\autoref{tab:ablation}, we can see that the smaller-sized pre-training corpus can drastically reduce the model's performance, with up to a 5\% drop seen on TriviaQA.
\begin{table}[!thb]
\centering
\small
\begin{tabular}{@{}llll@{}}
\toprule
Pre-train Examples         & NQ & TQA & WQ \\ \midrule
% \multicolumn{4}{c}{Pre-trained w/ DPR-Reverse  Corpus} \\
120K     & 42.5             & 48.2     & 39.7         \\
\midrule
% \multicolumn{4}{c}{Pre-trained  w/ Generated Corpus} \\ 
1M  & 42.8             & 48.8     & 40.2         \\ 
5M   & 43.8             & 51.5     & 41.7         \\
10M  & 44.3             & 52.1     & 42.5         \\
20M  & 44.5             & 53.2     & 43.0         \\
\bottomrule
\end{tabular}
\caption{Impact of pre-training corpus size on final downstream EM performance. The upper portion is pre-trained using the DPR-reverse corpus described in \autoref{sec:ablations} and the lower portion uses subsets of our generated pre-training corpus (\autoref{sec:pretraining_corpus})}
\label{tab:ablation}
\end{table}

\paragraph{Size of Memory}
\begin{figure}[!hb]
    \centering
    \includegraphics[width=0.94\linewidth]{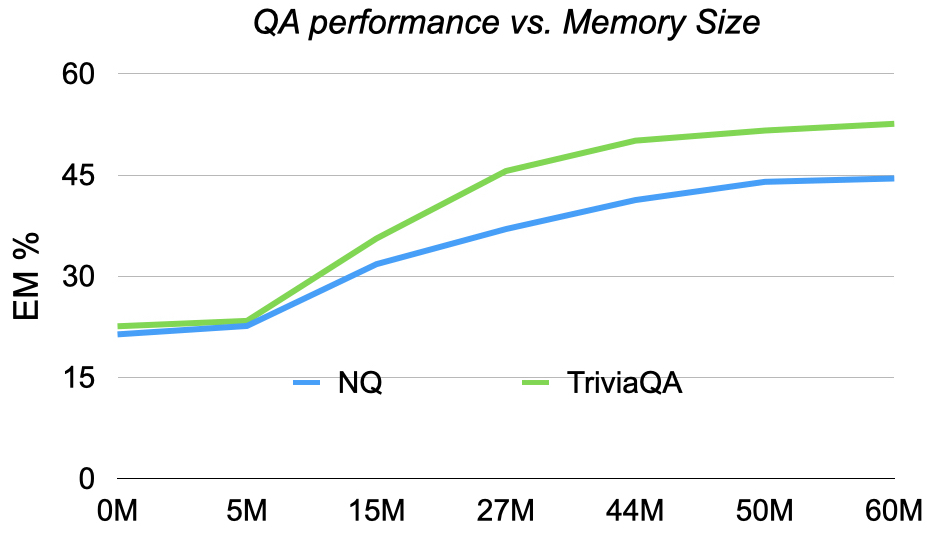}
    \caption{The impact of memory size on downstream QA EM performance.}
    \label{fig:memory_size}
\end{figure}

\begin{figure*}[!tb]
    \centering
    \includegraphics[width=1.0\linewidth]{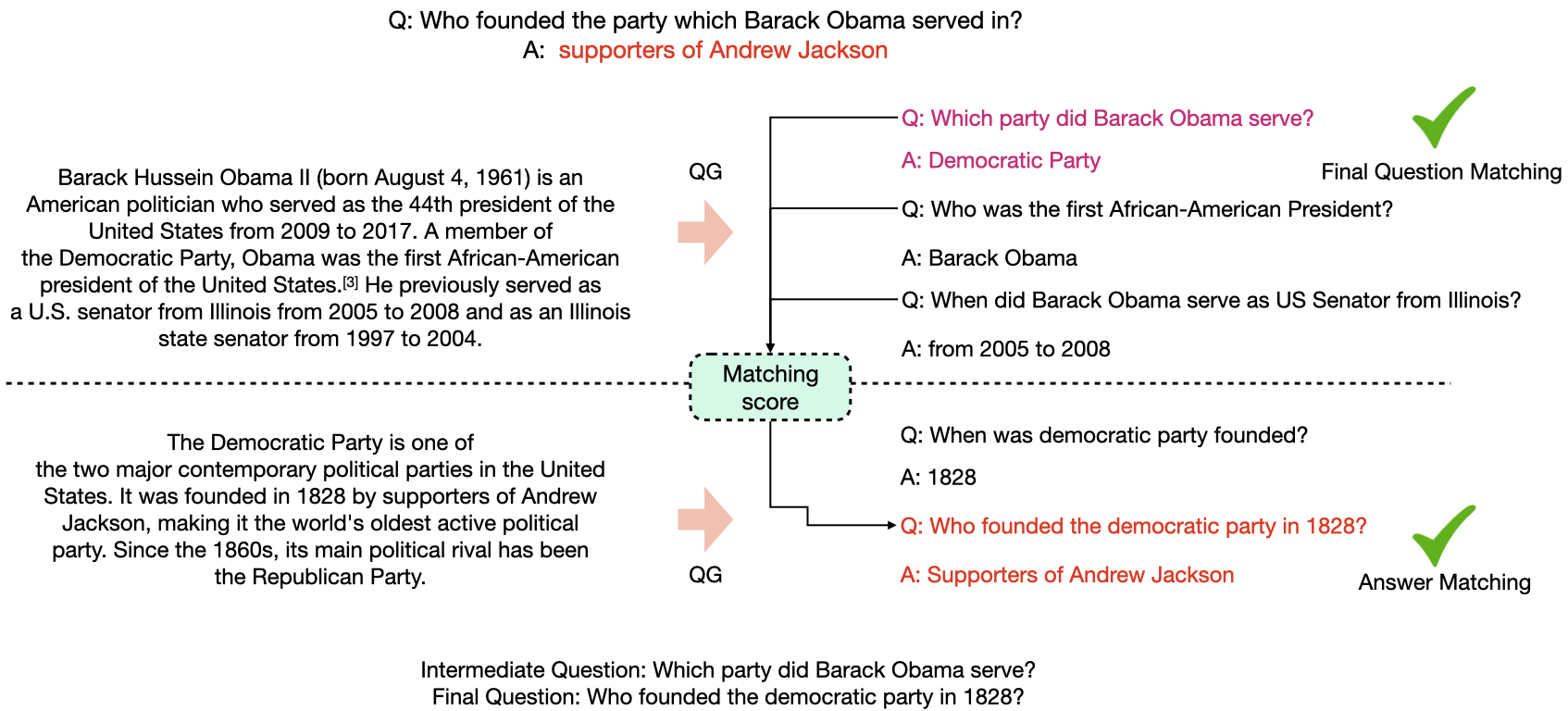}
    \caption{We first find the final question based on answer string matching with the pre-generated question, and then base on that to trace back the intermediate question.}
    \label{fig:mining}
\end{figure*}

Finally, we look at how big of memory we need to reach optimal downstream accuracy and how the model behaves with a smaller memory. As is shown in~\autoref{fig:memory_size}, having a small memory of less than 5M entries does not improve over a model with no memory at all. Due to the lack of coverage, the model does not receive a useful signal from the retrieval and is subsequently not incentivized to utilize those retrievals when making a prediction. However, once the size of the memory increases beyond 15M we observe a steep increase in the final performance, indicating that the model is gradually learning to incorporate retrieved information retrievals to assist prediction.

\section{MultiHop QA Training}
In order to train the multi-hop QA model, we need to have intermediate supervision for the query augmentation process. Here we use a string-based match to derive what are the most possible intermediate questions from a collection of pre-generated QA pairs. We depict the mining process as~\autoref{fig:mining}.

\end{document}